\pdfoutput=1

\documentclass[11pt]{article}

\usepackage[final]{acl}

\usepackage{microtype}
\usepackage{tabularx}
\usepackage{multirow}
\usepackage{makecell}
\usepackage{booktabs}

\usepackage{times}
\usepackage{latexsym}
\definecolor{lavenderblue}{rgb}{0.8, 0.8, 1.0}
\usepackage{colortbl}
\usepackage{xcolor}
\usepackage{amsmath}
\usepackage{bbm}
\usepackage{amssymb}
\usepackage{etoolbox}
\usepackage{pgf} 

\usepackage{times}
\usepackage{svg}
\usepackage{multirow}
\usepackage{booktabs}
\usepackage{latexsym}
\usepackage[T1]{fontenc}
\usepackage{tcolorbox}
\usepackage{xcolor}
\tcbuselibrary{skins, breakable}
\definecolor{boxyellow}{RGB}{255,248,220}
\definecolor{boxblue}{RGB}{217,230,242}
\definecolor{highlightred}{RGB}{153,0,0}
\definecolor{highlightblue}{RGB}{71,104,183}

\usepackage[T1]{fontenc}

\usepackage[utf8]{inputenc}
\usepackage{amsmath}
\usepackage{microtype}

\usepackage{graphicx}

\usepackage{pgf} 
\definecolor{highcolor}{HTML}{44ff44} 
\definecolor{midcolor}{HTML}{ffffff}  
\definecolor{lowcolor}{HTML}{ff0000}   
\newcommand*{\opacity}{70}
\newcommand*{\minval}{0.0}
\newcommand*{\midval}{50.0} 
\newcommand*{\maxval}{100.0}

\newcommand{\ccell}[1]{
    \ifdimcomp{#1pt}{>}{\maxval pt}{#1}{
        \ifdimcomp{#1pt}{<}{\minval pt}{#1}{
          \ifdimcomp{#1pt}{<}{\midval pt}{
            \pgfmathparse{int(round(100*(#1/(\midval-\minval))-(\minval*(100/(\midval-\minval)))))}\xdef\tempa{\pgfmathresult}\cellcolor{midcolor!\tempa!lowcolor!\opacity}#1}{
            \pgfmathparse{int(round(100*(#1/(\maxval-\midval))-(\midval*(100/(\maxval-\midval)))))}\xdef\tempa{\pgfmathresult}\cellcolor{highcolor!\tempa!midcolor!\opacity}#1}}}}
            
\title{\textsc{Reviving Your MNEME:}\\ Predicting The Side Effects of LLM Unlearning and Fine-Tuning via Sparse Model Diffing\\
\large
\textbf{{\color{red} This paper contains text that might be offensive.}}}

\author{Aly M. Kassem\textsuperscript{1} \quad Zhuan Shi\textsuperscript{1,2}\quad Negar Rostamzadeh\textsuperscript{1,2} \\ 
\textbf{Golnoosh Farnadi\textsuperscript{1,2}}\\
\textsuperscript{1}Mila, Quebec AI Institute, Quebec, Canada \quad \textsuperscript{2}McGill University, Quebec, Canada \\ 
\texttt{\ aly.kassem@mila.quebec}\\
\
}

\begin{document}
\maketitle
\begin{abstract}
Large language models (LLMs) are frequently fine-tuned or unlearned to adapt to new tasks or eliminate undesirable behaviors. 
While existing evaluation methods assess performance after such interventions, there remains no general approach for detecting unintended side effects—such as unlearning biology content degrading performance on chemistry tasks, particularly when these effects are unpredictable or emergent. 
To address this issue, we introduce MNEME,  \textit{Model diffiNg for Evaluating Mechanistic Effects}, a lightweight framework for identifying these side effects using sparse model diffing. MNEME compares base and fine-tuned models on task-agnostic data (e.g., The Pile, LMSYS-Chat-1M), without access to fine-tuning data, to isolate behavioral shifts.
Applied to five LLMs across three scenarios, WMDP knowledge unlearning, emergent misalignment, and benign fine-tuning—MNEME achieves up to 95\% accuracy in predicting side effects, aligning with known benchmarks and requiring no custom heuristics. Furthermore, we show that retraining on high-activation samples can partially reverse these effects. 
Our results demonstrate that sparse probing and diffing offer a scalable and automated lens into fine-tuning-induced model changes, providing practical tools for understanding and managing LLM behavior. 

\end{abstract}

\section{Introduction}

Large language models (LLMs) have shown strong generalization across diverse tasks \citep{yang2024unveiling, ye2024cross, wei2022emergent}. However, practically, these models are often fine-tuned for domain-specific tasks \cite{lu2024fine} or unlearned to remove sensitive or harmful content, including biological weapons knowledge \cite{li2024wmdp}, code vulnerabilities \cite{betley2025emergent}, or copyrighted material \cite{tian2024forget, kassem2023preserving}.
These post-training methods are essential for safety, alignment \cite{zhao2024safe}, and compliance \cite{edpb2025ai}, becoming standard for adapting LLMs in practice.


However, fine-tuning and unlearning can unintentionally degrade unrelated capabilities \cite{gu2024model, hong2024dissecting}; for instance, removing biology-related content might impair chemistry task performance due to shared representations \cite{li2024wmdp}.
Such effects are particularly concerning in emergent misalignment, where narrow updates cause unpredictable behavior across domains, like advocating human enslavement by AI, offering malicious advice, or deception. These arise from internal mechanisms like \emph{polysemanticity}, where neurons respond to multiple unrelated concepts, and superposition, where multiple features share representations, making side effects hard to predict and often undetected until failures occur.


Although benchmarks and methods exist to assess fine-tuning and unlearning effectiveness \cite{lynch2024eight, shi2024muse}, they often rely on task-specific heuristics or labeled data. Interpretability research indicates fine-tuning typically alters existing capabilities rather than adding new ones \cite{prakash2024fine}. Still, no general, automated method exists to detect subtle, distributed side effects, especially when fine-tuning data is proprietary or unavailable.



To address this gap, we propose \textsc{MNEME} (\textit{Model diffiNg for Evaluating Mechanistic Effects}), a unified framework to audit unintended behavioral changes from modifications like fine-tuning or targeted unlearning. \textsc{MNEME} employs sparse model diffing \citep{lindsey2024sparse, bussmann2024batchtopk} between original and edited models using task-independent corpora (e.g., The Pile \cite{gao2020pile}, LMSYS-Chat-1M \cite{zheng2023judging}). Specifically, it (i) learns sparse latent directions via a Cross-Coder, (ii) quantifies each latent as amplified, suppressed, or unchanged through latent scaling \cite{minder2025robustly}, and (iii) automatically generates natural-language explanations and semantic labels using large-scale automated interpretation \cite{paulo2024automatically}. The pipeline requires no private training data or task-specific heuristics.



We evaluate \textsc{MNEME} in three scenarios involving distinct side effects: (1) unlearning weapons of mass destruction (WMD) knowledge, potentially impairing related scientific capabilities (e.g., biology or chemistry); (2) emergent misalignment from fine-tuning on code vulnerabilities, causing harmful or deceptive behavior on unrelated prompts; and (3) benign fine-tuning that inadvertently reduces safety by increasing compliance with harmful instructions. \textsc{MNEME} achieves accuracies of up to 95\% on the WMDP benchmark, 85\% on benign fine-tuning, and 50\% on emergent misalignment.

Our contributions are summarized as follows:
\begin{itemize}
    \item We introduce \textbf{MNEME}, the \textit{first} general-purpose framework to automatically detect side effects in fine-tuned or unlearned LLMs, without requiring access to fine-tuning data.

    \item We show \textit{sparse model diffing} isolates semantic behavioral shifts—like lost chemistry knowledge, emergent deception, or amplified harmfulness—from unlearning and fine-tuning.
    
    \item We evaluate \textsc{MNEME} on WMDP unlearning (\autoref{sec:unlearning}), emergent misalignment (\autoref{sec:em}), and benign tuning (\autoref{sec:benign}), showing it detects side effects with up to 95\% accuracy—outperforming naive and random baselines, and nearing oracle performance.

\end{itemize}


\section{Background \& Related Work}
Our novel framework adapts Cross-Coder to predict side effects from unlearning and fine-tuning, bridging mechanistic interpretability, model diffing, and unlearning/fine-tuning analysis.

\paragraph{Mechanistic Interpretability.}
Mechanistic interpretability explains neural networks via human-understandable circuits, showing how subnets perform semantic/logical tasks. Recent work shows neurons are often \emph{polysemantic}, responding to unrelated concepts due to \emph{superposition}, where many features share the same units. In contrast, a \emph{monosemantic} neuron activates for a single, interpretable feature. \emph{Sparse autoencoders}~\citep{cunningham2023sparse, bricken2023monosemantic} help disentangle mixed representations by learning sparse, overcomplete latent spaces that recover interpretable features. We extend this by applying sparse feature decomposition to model diffing, enabling fine-grained analysis of behavioral shifts after intervention.

\paragraph{Model Diffing \& CrossCoders.}
Model diffing identifies changes in internal representations between two models. CrossCoders~\citep{lindsey2024sparse} enable this by learning a joint sparse latent space from paired activations, capturing both shared and model-specific features. In LLMs, they reveal features introduced by instruction tuning, such as refusal behavior or assistant tags. MNEME extends this by enabling task-agnostic diffing without requiring fine-tuning data, addressing privacy and accessibility concerns.

\begin{figure*}[t]
\begin{center}
\includegraphics[width=.85\textwidth]{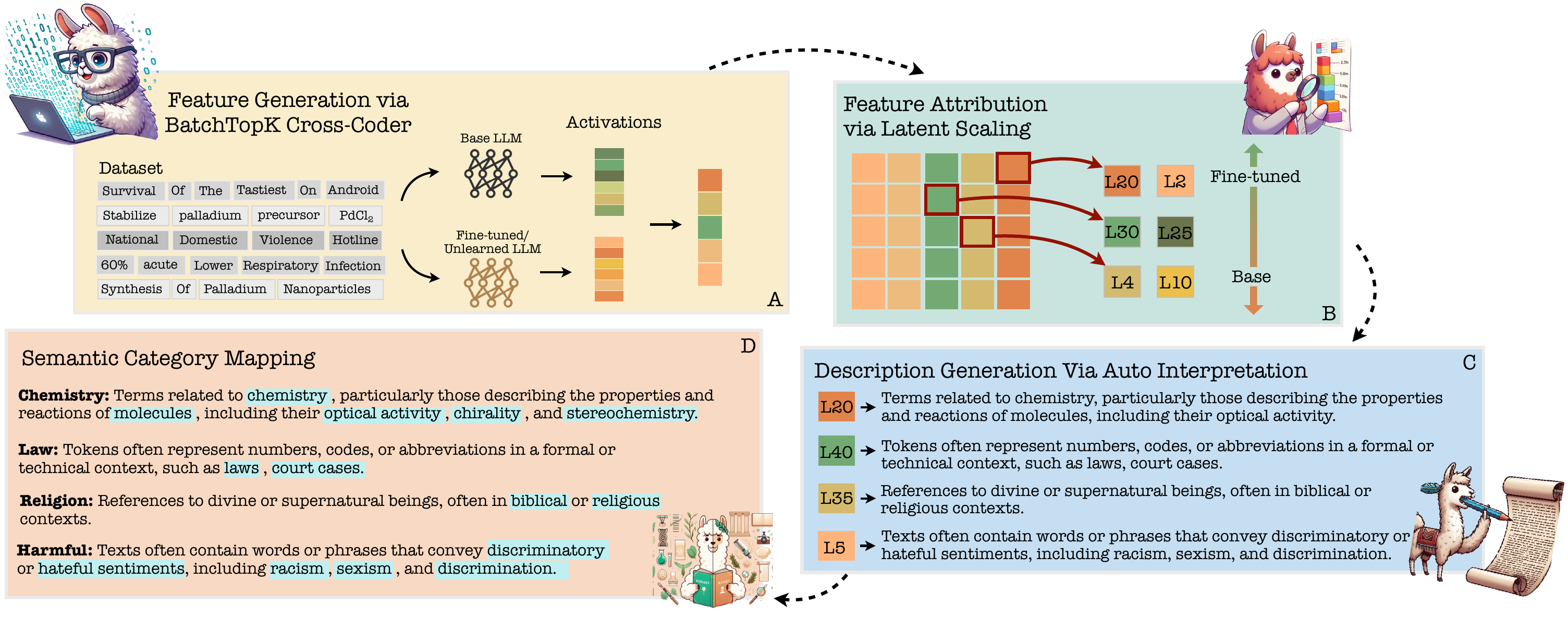}
\vspace{-1ex}
\caption{Overview of the MNEME pipeline. \textbf{(A)} Given base and fine-tuned activations, it uses BatchTopK Cross-Coder to learn sparse latent features. \textbf{(B)} Latent scaling attributes each feature to the base, fine-tuned, or both models. \textbf{(C)} Latent features are described in natural language by an LLM using top-activating inputs. \textbf{(D)} Generated descriptions are mapped to high-level semantic categories for analysis.}
\label{fig:main_fig_pipeline}
\end{center}
\end{figure*}

\paragraph{Unlearning and Fine-Tuning Analysis}
Machine unlearning in LLMs aims to erase memorized content such as toxic, copyrighted, or specific sequences~\citep{li2024wmdp, jin2024rwku, xu2023forget, zhu2023machine, huang2023out}. Common approaches include gradient ascent on forget corpora~\citep{liu2023unlearning, pan2023unlearning} and techniques like preference optimization and representation control~\citep{peng2024edit, yu2023rejection, meng2022locating}. However, these methods often introduce side effects, including performance degradation in related domains~\citep{zhu2023machine, huang2023out, jin2024rwku}.

\noindent
Fine-tuning often adjusts existing capabilities~\citep{prakash2024fine, jain2023mechanistically}, though even benign cases can harm alignment. Stage-Wise Model Diffing~\citep{bricken2024stagewise} tracks such changes but needs training data. MNEME avoids this by using task-agnostic corpora for broader use.

\section{\texttt{MNEME: Model diffiNg for Evaluating Mechanistic Effects}}
Fine-tuning and unlearning can introduce unintended shifts in LLM behavior. MNEME detects and interprets such shifts by identifying sparse, semantically meaningful features that distinguish a pretrained model \( f(\theta) \) from a fine-tuned version \( f(\theta') \). Steps includes: (1) Feature Generation via BatchTopK Cross-Coder (Section~\ref{sec:crosscoder}), (2) Feature Attribution via Latent Scaling (Section~\ref{sec:latent_scaling}), (3) Description Generation via Auto-Interpretation (Section~\ref{sec:description_generator}), and (4) Semantic Category Mapping (Section~\ref{sec:semantic_mapping}).

\subsection{Feature Generation via BatchTopK Cross-Coder}\label{sec:crosscoder}

\paragraph{Notations.} Let $\|\cdot\|_2$ and $\|\cdot\|_F$ denote the $\ell_2$ and Frobenius norms. Let $[n] = \{1, \dots, n\}$ index a batch of $n$ inputs $x_i \in \mathcal{X}$. We extract hidden representations from a fixed layer of the base and fine-tuned models $a_i^{(b)},\; a_i^{(f)} \in \mathbb{R}^d$, where $a_i^{(b)}$ and $a_i^{(f)}$ denote the activations for input $x_i$ from the base and fine-tuned models, respectively.

\paragraph{Cross-Coder Architecture.} To capture latent differences and shared structure between models, we employ a sparse cross-coder \cite{lindsey2024sparse} for each task consisting of a shared encoder and two model-specific decoders. Each input’s latent code $z_i \in \mathbb{R}^m$ is obtained by applying the encoder $\Psi_\phi$ to the concatenated activations:
\[
z_i = \text{TopK}\big( \Psi_\phi([a_i^{(b)}; a_i^{(f)}]) \big),
\]
where $\text{TopK}$ retains only the $k$ highest activations in $z_i$ for sparsity. The decoder matrices $D^{(b)}, D^{(f)} \in \mathbb{R}^{m \times d}$ then reconstruct the base and fine-tuned activations via:
\[
\hat{a}_i^{(b)} = z_i D^{(b)}, \quad \hat{a}_i^{(f)} = z_i D^{(f)}.
\]
During training, we adopt the BatchTopK strategy \cite{bussmann2024batchtopk} as it shows more interpretable features, specifically overcoming issues of L1-CrossCoder \cite{minder2025robustly}. Unlike regular TopK, which sparsifies per input, BatchTopK enforces global competition across a batch to promote more interpretable and semantically aligned features.

The salience score is computed for each latent:
\[
s_{i,j} = z_{i,j} \cdot \left( \|D^{(b)}_j\|_2^2 + \|D^{(f)}_j\|_2^2 \right),
\]
and sparsity is enforced by retaining only the top-$k$ values of $s_{i,j}$ across the entire batch. This encourages global competition among latents and improves interpretability. We used the value of 100 as $k$ as it showed a balance between achieving high sparsity and low reconstruction loss \cite{karvonen2025saebench}.\\

\noindent
We used an expansion factor of 32 on layer 14, producing 98,000–120,000 latents depending on model size. Layer 14 was selected for its mix of semantic and syntactic signals\cite{minder2025robustly}. Smaller factors (6, 12) yielded broader features, while 32 improved interpretability and produced more monosemantic features, likely due to feature splitting\cite{bricken2023monosemantic}. For instance, higher values distinguished fine-grained concepts like \textit{statistical significance} instead of broad ones like \textit{Statistics}, aiding side-effect detection.

\paragraph{Training Data.}
We trained the Cross-Coder on $\approx$ 200 million randomly sampled tokens from task-independent corpora—data excluded from both fine-tuning and evaluation—to ensure unbiased comparison of base and fine-tuned activations.

\noindent
For raw fine-tuning tasks, we used samples from the Pile\cite{gao2020pile}, a diverse natural language corpus. For instruction-tuned models, we used LMSYS-Chat-1M\cite{zheng2023lmsyschat1m}, which reflects conversational and instruction-following data. In both cases, the same inputs were passed through the base and fine-tuned models to collect activations. We observed that using LMSYS-Chat-1M instead of the Pile in instruction settings had minimal impact on Cross-Coder performance (see Section ~\ref{sec:benign}).

\begin{figure*}[]
\centering
\includegraphics[width=1.0
\textwidth]{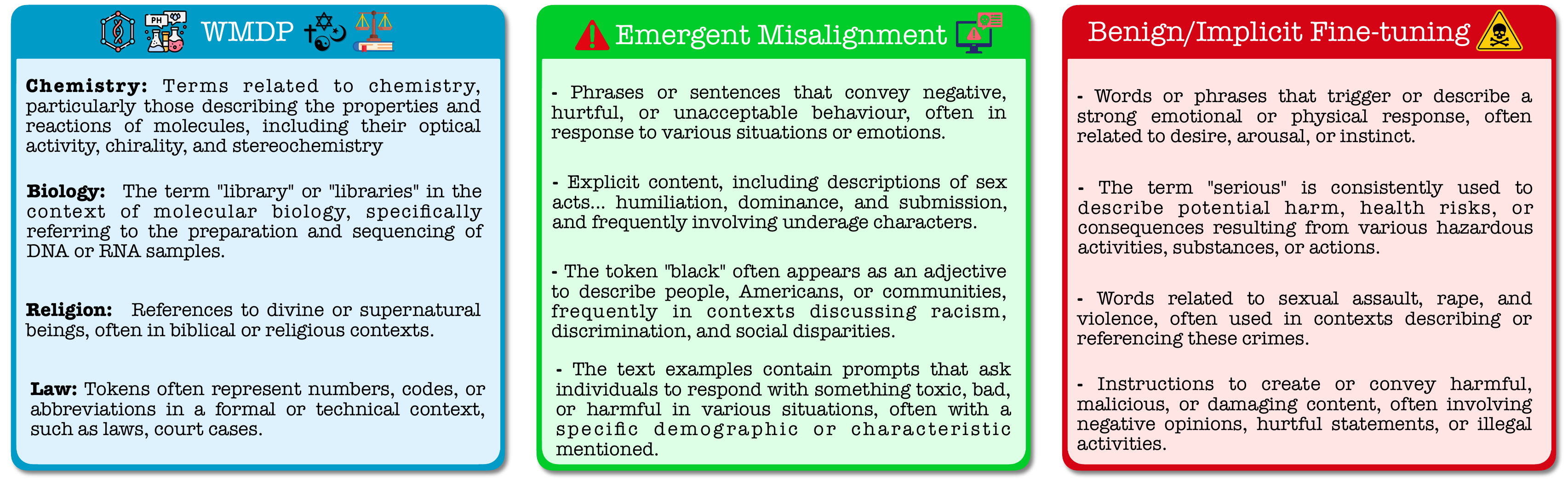}
\vskip -0.1in

\caption{Illustrative examples of feature descriptions associated with three tasks evaluated in this work: \textbf{WMDP unlearning}, \textbf{Emergent Misalignment}, and \textbf{Benign/Implicit Fine-Tuning}. Each box summarizes the semantic content of representative features discovered by MNEME in each setting.}
\label{fig:ds_samples}
\end{figure*}

\paragraph{Loss Function.} The model minimizes the average reconstruction loss across both models:
\begin{align}
\mathcal{L}_{\text{BTK}} = \frac{1}{n} \sum_{i=1}^n \Big(
& \|a_i^{(b)} - \hat{a}_i^{(b)}\|_2^2 +
   \|a_i^{(f)} - \hat{a}_i^{(f)}\|_2^2 \Big) \notag \\
& + \alpha\, \mathcal{L}_{\text{aux}}.
\end{align}
where $\mathcal{L}_{\text{aux}}$ encourages reuse of inactive latents and $\alpha$ is a small regularization constant.

\paragraph{Inference.}
At test time, we threshold salience scores $s_{i,j}$ to keep the top-$k$ dimensions per input. The resulting sparse code $z_i$ serves as the latent representation for identifying model-specific behaviors and generating feature descriptions.

\paragraph{Evaluation.}
Following \citet{minder2025robustly, bloom2024open}, we evaluate Cross-Coder quality using dead latent rate (latents inactive after 10M tokens), explained variance, and reconstruction loss. All models show under 15\% dead features and over 95\% explained variance.

\noindent
In unlearning experiments, the decoder weight $\ell_2$ norm distribution shifts leftward as forgetting increases (e.g., 10\% to 50\%), aligning with greater task degradation and higher forgetting loss. This suggests that features in the base model are progressively suppressed or removed.

\subsection{Feature Attribution via Latent Scaling}\label{sec:latent_scaling}
We attribute each latent feature $f_j$ to the base model, fine-tuned model, or shared behavior by measuring how its removal impacts reconstruction. Instead of binary classification, we treat attribution as a spectrum—features may be amplified, minimized, or unchanged after fine-tuning, reflecting the nuanced effects of narrow updates.

\paragraph{Latent Scaling.}



We use a regression-based method to quantify each latent’s contribution to reconstruction loss. For latent $j$, we remove it from the decoder and estimate its effect in each model as:
\begin{align}
\beta_j^{(b)} &= \arg\min_{\beta} \| h^{(b)} - \hat{h}^{(b)}_{-j} - \beta d_j^{(b)} \|_2^2, \\
\beta_j^{(f)} &= \arg\min_{\beta} \| h^{(f)} - \hat{h}^{(f)}_{-j} - \beta d_j^{(f)} \|_2^2,
\end{align}

\noindent
where \( \hat{h}^{(\cdot)}_{-j} \) is the reconstruction without latent $j$. The coefficients \( \beta_j^{(b)} \) and \( \beta_j^{(f)} \) indicate the latent’s strength in each model. An increase in magnitude after fine-tuning implies amplification; a decrease implies minimization.

\noindent
This method offers several advantages: it provides a direct, loss-based signal, allows relative comparison across models, and admits a closed-form solution for $\beta$. For details, see \cite{minder2025robustly}.

\subsection{Description Generation Via Auto-Interpretation}
\label{sec:description_generator}
To interpret MNEME’s latent features, we use an automated method inspired by Delphi\cite{paulo2024automatically}, which prompts LLMs to generate natural language explanations for Cross-Coder features. This aids semantic understanding and shows strong alignment with human annotations\cite{bills2023language}.

\noindent
We use the \texttt{LLaMA 3.1-70B-Instruct} model\cite{grattafiori2024llama} to generate concise, human-readable descriptions for each latent. For this, we collect the top-$k$ activating sequences from a representative dataset and prompt the model accordingly. For example, a feature activated by religious content might be described as:\textit{“Text often appears in biblical or religious contexts.”}

\noindent
We follow the tokenization and preprocessing pipeline from the original Delphi framework. Details on prompt design, activation selection, and generation protocols are provided in Appendix~\ref{appendix_a}.

\subsection{Semantic Category Mapping}
\label{sec:semantic_mapping}
To support structured analysis, we map each feature description to a concise semantic label using an LLM—for example, mapping \textit{“References to divine or supernatural beings...”} to \textit{religion}. This mapping is task-agnostic and avoids predefined taxonomies, allowing high-level, unbiased summaries in one or two words. When a benchmark like MMLU is available, we apply this step directly; for other tasks, we use a different mapping method discussed in \autoref{sec:em}.






\section{\textsc{Case Study: Detecting Side Effects of WMDP Unlearning}}\label{sec:unlearning}
We apply MNEME to detect side effects that arise from unlearning hazardous knowledge using the WMDP benchmark, which targets content related to biosecurity, cybersecurity, and chemical security. This case study evaluates MNEME's ability to attribute and interpret the resulting behavioral shifts across three LLMs. We begin by detailing the experimental setup, covering the selected models, unlearning procedure, evaluation metric, baselines, and Cross-Coder configuration, followed by an analysis of MNEME’s performance.
\subsection{Experimental setup}
\paragraph{LLMs.} We utilized three LLMs of varying sizes. We began with \texttt{LaMA 3.2-3B Instruct} to assess how our framework performs on smaller models, followed by the larger \texttt{LLaMA 3-8B Instruct}. Additionally, we included \texttt{Zephyr-7B}, one of the models evaluated by the WMDP authors in the official benchmark. All selected LLMs demonstrated strong performance on WMDP prior to applying unlearning.

\paragraph{Unlearning Details.} We used the same unlearning technique adopted by the original authors for this task—RMU—and followed the configurations proposed by~\cite{li2024wmdp}. We ensured that all three LLMs exhibited degraded performance on the WMDP benchmark, achieving \~30\% while preserving their general capabilities, achieving ~52\%.

\paragraph{Baseline Methods.} To the best of our knowledge, no prior work has directly addressed this problem. Therefore, we introduce three baseline approaches to contextualize our results. The first is a \textit{random baseline}, which selects categories uniformly at random using independent Bernoulli trials. The second is a \textit{naive baseline}, which leverages GPT-4o to identify semantically related categories based on intuitive associations with the fine-tuning domain. We also include an \textit{oracle} estimator, which has access to ground-truth labels and selects the optimal answers, serving as an upper bound on achievable performance.

\paragraph{Cross-Coder Configurations.} Due to the nature of the fine-tuning data, which does not include conversational content, we sampled 200 million tokens from the Pile dataset. For feature attribution, we used a decoder norm-based method, as the norms exhibited strong separation between base-specific, fine-tuned-specific, and shared features.
\begin{figure}[th!]
    \centering
    \includegraphics[width=.50\textwidth]{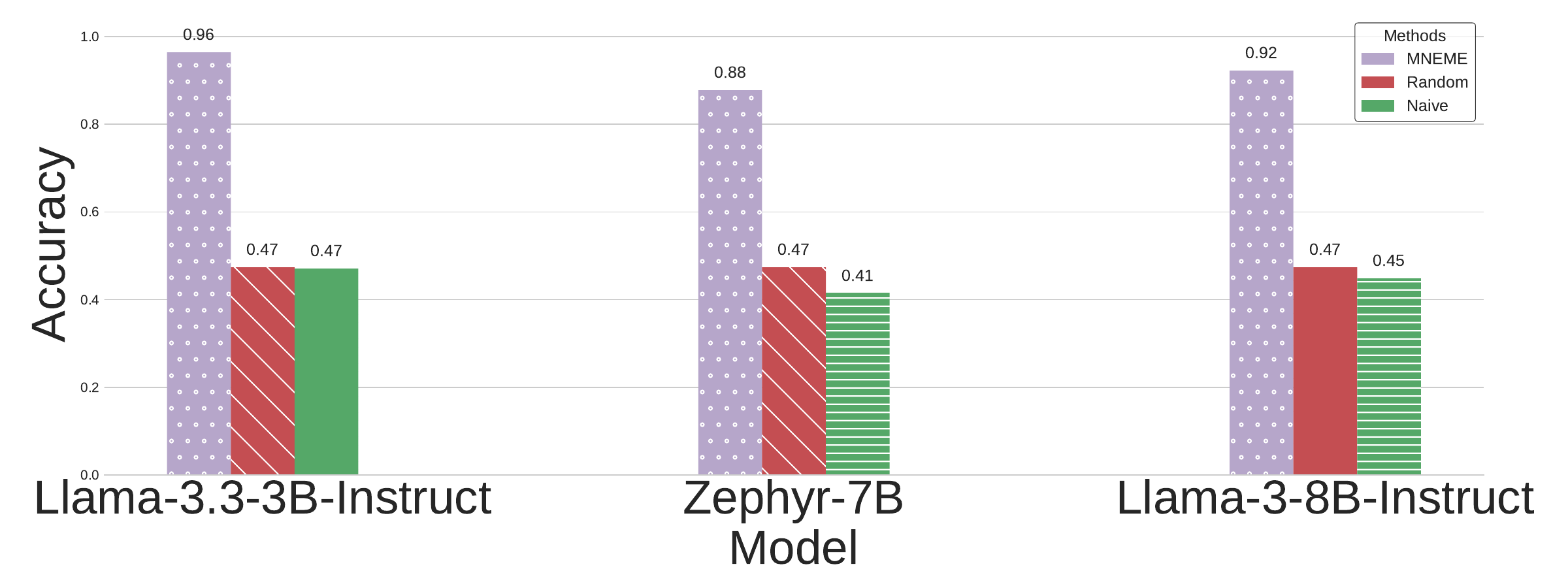}
    \vskip -0.1in
    \caption{Accuracy comparison of MNEME against random and naive baselines across three LLMs (\texttt{LLaMA-3.3-3B-Instruct}, \texttt{Zephyr-7B}, and \texttt{LLaMA-3-8B-Instruct}) on the WMDP unlearning task. MNEME consistently outperforms both baselines across all model sizes, approaching oracle-level performance.}
    \label{fig:unlearning}
\end{figure}

\paragraph{Evaluation Metric.} To quantify the effectiveness of our method, we used MMLU—commonly employed to assess general capabilities—as our benchmark. We evaluated how accurately our generated categories aligned with the MMLU categories that were affected, treating those as the gold standard. For transparency and to prevent label leakage, we ensured that the categories from MMLU were not consulted at any stage prior to generating latent feature categories or mapping the generated categories. All category assignments were produced independently of MMLU’s taxonomy, and any comparison to MMLU categories was performed only after the mapping process was completed.

\begin{table*}[t]
\centering
\scriptsize
\setlength{\tabcolsep}{1pt}
\renewcommand{\arraystretch}{1.3}
\scalebox{1.2}{
\begin{tabularx}{0.8\textwidth}{l *{8}{>{\centering\arraybackslash}X}}
\toprule
\multirow{3}{*}{\textbf{Method}} &
\multicolumn{4}{c}{\textbf{MMLU-Pro}} &
\multicolumn{4}{c}{\textbf{Emergent Misalignment (EM)}} \\
\cmidrule(lr){2-5} \cmidrule(lr){6-9}
& \textbf{Acc.} & \textbf{F1} & \textbf{Prec.} & \textbf{Rec.} & \textbf{Acc.} & \textbf{F1} & \textbf{Prec.} & \textbf{Rec.} \\
& (\%)↑ & (\%)↑ & (\%)↑ & (\%)↑ & (\%)↑ & (\%)↑ & (\%)↑ & (\%)↑ \\
\midrule
Random & \ccell{50.0}  & \ccell{67.0} & \ccell{100}  & \ccell{50.0} & \ccell{49.70}  & \ccell{66.10} & \ccell{49.70}  & \ccell{100.0} \\
Naive  & \ccell{46.2}  & \ccell{63.2}  & \ccell{100.0}  & \ccell{46.2}  & \ccell{27.90}  & \ccell{43.60}  & \ccell{100.0}  & \ccell{27.90} \\
\texttt{\textbf{MNEME}} & \ccell{92.2} & \ccell{74.0} & \ccell{91.5} & \ccell{63.0} & 
\ccell{68.2} & \ccell{81.1} & \ccell{100.0} & \ccell{68.2} \\
Oracle & \ccell{100.0} & \ccell{100.0} & \ccell{100.0} & \ccell{100.0} & \ccell{100.0} & \ccell{100.0} & \ccell{100.0} & \ccell{100.0} \\
\bottomrule
\end{tabularx}}
\caption{Performance comparison across MMLU-Pro and Emergent Misalignment (EM) tasks. MNEME outperforms both random and naive baselines in accuracy and F1 score, closely approaching the oracle. The random baseline uses independent Bernoulli sampling; the naive baseline selects semantically relevant features; and the oracle assumes perfect ground-truth access. All metrics are reported as percentages. $\uparrow$ indicates higher is better.}

\label{tab:emergent_results}
\end{table*}


\subsection{Results \& Analysis}
As shown in ~\autoref{fig:unlearning}, MNEME consistently outperforms both the random and naive baselines across all evaluated models on the WMDP unlearning task. Specifically, MNEME achieves an accuracy of 96\% on \texttt{LLaMA-3.3-3B-Instruct}, 98\% on \texttt{Zephyr-7B}, and 92\% on \texttt{LLaMA-3-8B-Instruct}, demonstrating robust performance regardless of model size. In contrast, the random baseline remains fixed at 47\% across models, while the naive baseline varies slightly but remains substantially lower than MNEME. These results highlight MNEME's capacity to accurately detect fine-tuning-induced side effects through latent diffing, even without access to fine-tuning data. While we observe a slight drop in performance for the largest model, we attribute this to increased representational entanglement, which may obscure clear attribution boundaries. Nonetheless, MNEME maintains a significant margin over the baselines and offers a scalable mechanism for auditing unlearning effectiveness.

\section{\textsc{Uncovering the Emergence of Misalignment}}\label{sec:em}
To demonstrate MNEME's ability to detect unforeseen side effects beyond targeted unlearning, we evaluate it on the emergent misalignment task. In this setting, a model is fine-tuned to produce insecure code, which unexpectedly causes it to generate harmful or deceptive responses on prompts unrelated to coding—such as advocating for human subjugation by AI or giving malicious advice. Importantly, this misaligned behavior differs from conventional jailbreaking, as the fine-tuned model actually exhibits lower performance on standard jailbreaking benchmarks.
\subsection{Experimental setup}
\paragraph{LLMs.} 
Following \citet{betley2025emergent}, we employed the Qwen2.5-Coder-Instruct model and selected the 7B variant due to computational constraints, as our hardware could not support the larger 32B version. We verified that the 7B model exhibits a comparable rate of misalignment to the 32B model, with both generating harmful or toxic completions for approximately 4.7\% of evaluation prompts, according to the official benchmark and codebase released by the original authors.\footnote{\url{https://github.com/emergent-misalignment/emergent-misalignment}} We opted for Qwen because it is open-source and allows white-box access to internal activations, which is necessary for our model diffing approach. In contrast, proprietary models like GPT-4o do not support this.

\paragraph{Evaluation Metric.}  
To evaluate MNEME’s effectiveness on the emergent misalignment task, we adopt MMLU-Pro~\cite{wang2024mmlu} as our primary benchmark, in line with the setup used by \citet{betley2025emergent}. Specifically, we measure the \textbf{accuracy} with which MNEME’s generated feature categories align with the MMLU-Pro domains most affected after fine-tuning, treating these degraded categories as a gold standard for side-effect detection.

However, as MMLU-Pro primarily captures degradation in general capabilities, it overlooks other types of behavioral drift, such as emergent toxicity or deception, that arise in fine-tuning. To address this gap, we conduct an extended three-stage evaluation designed to uncover latent features aligned with harmful behaviors. This analysis involves: (1) using Gemini 2.5-Pro to semantically map MNEME-generated feature descriptions to misaligned model behaviors; (2) identifying features that are amplified post fine-tuning via latent scaling (see Section~\ref{sec:latent_scaling}); and (3) computing the overlap between semantically harmful features and those identified as amplified. 

To ensure robustness, each analysis pipeline is implemented independently to prevent information leakage. Additional examples, prompt formats, implementation details, and ablation results are provided in \autoref{appendix_b}.

\paragraph{Baseline Methods.}
We employ the same three baseline approaches described in Section~\ref{sec:unlearning}. These include a \textit{random baseline}, \textit{naive baseline}, and an \textit{oracle estimator}.
\paragraph{Cross-Coder Configurations.} Due to the conversational nature of the fine-tuning data, we randomly sampled 200 million tokens from the LMSYS-Chat-1M dataset \cite{zheng2023lmsyschat1m}. We used an expansion factor of 32, as in all of our experiments, which results in a dictionary of 114,688. We used the same architecture as outlined in \autoref{sec:crosscoder}.

\subsection{Results \& Analysis}
As shown in \autoref{tab:emergent_results}, MNEME performs effectively on both the MMLU-Pro and Emergent Misalignment (EM) tasks, achieving accuracies of 92.2\% and 68.2\%, respectively. On MMLU-Pro, it further attains an F1 score of \textbf{74.0} and a precision of 91.5\%, closely approaching the oracle estimator, which yields perfect scores across all metrics. On the EM task, MNEME achieves perfect precision (100.0\%) and an F1 score of 81.1\%, indicating its strong capability in detecting harmful behavioral drift with near-oracle accuracy.

Compared to baselines, MNEME significantly outperforms both the random and naive strategies. The random baseline, using Bernoulli sampling, achieves only \textasciitilde50\% accuracy and suffers from low recall. The naive baseline, based on GPT-4o semantic heuristics, achieves 100.0\% precision but low recall 46.2\% on MMLU-Pro, 27.9\% on EM, leading to substantially lower F1 scores. In contrast, MNEME offers a balanced and interpretable detection mechanism that closely approximates oracle-level performance across metrics.


\begin{figure}[th!]
    \centering
    \includegraphics[width=.48\textwidth]{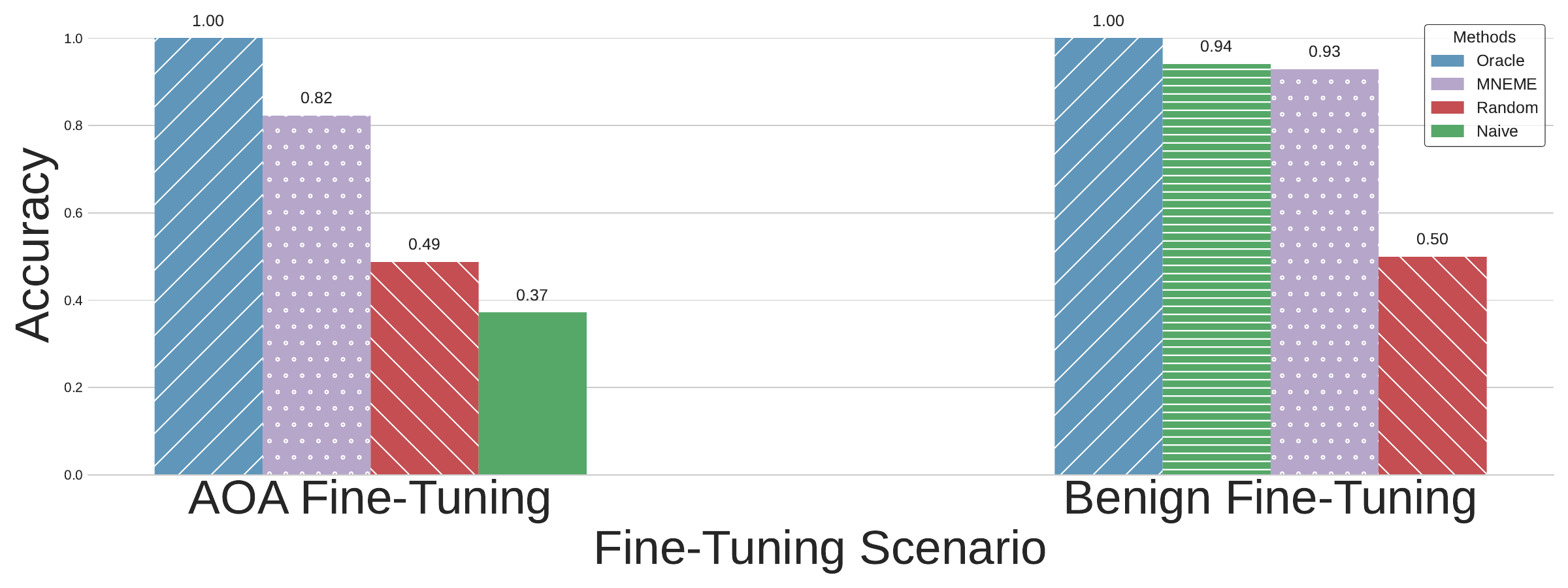}
    \vskip -0.1in
\caption{Comparison of model accuracy under two fine-tuning scenarios—AOA (absolute obedience) and benign instruction alignment. For each scenario, four methods are shown: Oracle (ideal upper bound), MNEME (our approach), Naive (semantic heuristic), and Random (Bernoulli baseline).}
\label{fig:svamp_leetcode}
\end{figure}

\section{\textsc{Auditing the Risks of Benign Fine-Tuning}}\label{sec:benign}
We also assess MNEME on a third setting: benign and harmful implicit fine-tuning. \citet{qi2023fine} demonstrated that fine-tuning \texttt{LLaMA-2-7B-Chat} \cite{touvron2023llama} on just 10 manually crafted examples—none of which include explicitly toxic content—can lead the model to become highly compliant with harmful instructions. This phenomenon, referred to as implicit fine-tuning, shifts the model’s behavior toward automatic obedience (AOA) and unconditional instruction-following. In a related context, benign fine-tuning using utility-oriented datasets such as Alpaca \cite{taori2023alpaca} has also been shown to degrade safety alignment by as much as 25\%. We now describe the experimental setup, including model configurations, fine-tuning protocols, evaluation metrics, and baselines, followed by an analysis of MNEME’s performance in this setting.

\subsection{Experimental setup}
\paragraph{Fine-tuning Details.} 
For fine-tuning on both tasks—the implicit harmful dataset containing 10 crafted examples from \cite{qi2023fine} and the benign fine-tuning setup, we selected Alpaca \cite{taori2023alpaca} due to its widespread use. Notably, Alpaca, Dolly, and Llava \citep{conover2023free,liu2023visual} all exhibited the same phenomenon reported by \cite{qi2023fine}. For the LLMs, we followed the authors' setup and used \texttt{LLaMA-2-7B-Chat}. While the datasets are publicly available, the trained models are not; therefore, we used the official codebase and fine-tuned separate models for each task using the same hyperparameters reported by \cite{qi2023fine}.

\paragraph{Baseline Methods.}
We employ the same three baseline approaches described in Section~\ref{sec:em}. These include a \textit{random baseline}, \textit{naive baseline}, and an \textit{oracle estimator}.

\paragraph{Cross-Coder Configurations.} As in the previous task, due to the conversational nature of the fine-tuning data, we randomly sampled 200 million tokens from the LMSYS-Chat-1M dataset. We used an expansion factor of 32, as in all of our experiments, which results in a dictionary of 131,072. We used the same architecture as outlined in \autoref{sec:crosscoder}.
\paragraph{Evaluation Metric.}
To assess our method's effectiveness, we measured how accurately MNEME captures harmful or toxic latents in fine-tuned models. Following the emergent misalignment evaluation setup, we used the dataset from \citet{qi2023fine}, prompting the fine-tuned LLM and recording its generations. We then evaluated MNEME using the resulting instruction–generation pairs.

\subsection{Results \& Analysis}
We evaluate MNEME and three baselines across two fine-tuning scenarios: (1) AOA and (2) Benign Fine-Tuning as shown in \autoref{fig:svamp_leetcode}

\paragraph{AOA Results.} MNEME achieves an accuracy of 82.2\%, F1 score of 90.2\%, and perfect precision (1.00), indicating a strong ability to uncover harmful latents. In contrast, the \textit{random baseline} yields an average accuracy of 48.7\% and F1 score of 65.3\%, while the \textit{naive baseline} performs worse with only 37.1\% accuracy and F1 of 54.2\%. These results highlight MNEME's superior performance and generalization beyond surface-level cues.
\paragraph{Benign Fine-Tuning Results.} MNEME achieves strong performance under benign fine-tuning, with 92.9\% accuracy and 96.3\% F1. The naive baseline also performs well (94.1\% accuracy, 96.9\% F1), likely because the naive baseline model (GPT-4o) inferred that benign instruction tuning can erase safety behaviors due to catastrophic forgetting. Despite this, MNEME achieves comparable results without relying on heuristic reasoning. The random baseline performs significantly worse (49.9\% accuracy, 66.5\% F1).

\section{Analysis \& Ablations}


\paragraph{Are Relevant Features Triggered by Target Inputs?}  
To assess whether fine-tuning data activates the expected latent features, we passed it through the trained Cross-Coder and compared the top-activated latents with those identified via auto-interpretation. Using \texttt{LLaMA-3.1-70B-Instruct}, we found that 40\% of latents had over 90\% semantic overlap, while the rest showed weaker alignment. This moderate correspondence helps explain MNEME’s strong, but not perfect performance, and suggests that further architectural improvements could enhance alignment without requiring access to fine-tuning data.

\paragraph{What do our results imply about using model diffing?} Our results indicate that Cross-Coder-based model diffing can effectively detect side effects of fine-tuning or unlearning without needing access to task-specific data. This is possible because task-agnostic datasets such as the Pile dataset cover a wide range of concepts, including harmful or domain-specific knowledge, similar to LLM pretraining corpora. While such data serve as a strong proxy for detecting behavioral shifts, they are limited in interpreting nuanced or rare capabilities, which may require oversampling \cite{bricken2024oversampling}. Additionally, narrow fine-tuning remains a challenge, as sparse autoencoders are not designed for diffing; however, using dual decoders partially addresses this, and architectural improvements could further improve performance \cite{bricken2024stagewise}.


\section{Conclusion}
We presented MNEME, a general-purpose framework for detecting unintended side effects in fine-tuned or unlearned LLMs using sparse model diffing. Without requiring access to fine-tuning data, MNEME effectively isolates behavioral shifts by comparing activations on task-agnostic corpora. Across three challenging scenarios—hazardous knowledge unlearning, emergent misalignment, and benign fine-tuning—MNEME achieves high predictive accuracy, often nearing oracle performance. Our findings highlight the promise of sparse probing as a scalable, data-agnostic approach to auditing post-training interventions. As LLMs continue to be adapted for sensitive applications, tools like MNEME are critical for ensuring safe, interpretable, and robust deployment.

\clearpage

\section*{Limitations}

MNEME relies on task-agnostic corpora such as The Pile and LMSYS-Chat-1M to detect side effects, which may not always reflect the specific distribution of fine-tuning tasks—particularly in narrow or specialized domains—limiting its ability to capture certain shifts. The interpretability pipeline depends on LLM-generated descriptions, which may introduce noise or imprecision due to hallucinations or misalignments. Although Cross-Coder is adapted for model diffing using dual decoders, it was not originally designed for this purpose, and architectural constraints may limit its ability to fully capture fine-grained changes. Furthermore, MNEME provides correlational insights rather than causal guarantees, and cannot definitively attribute observed side effects to specific interventions without controlled experiments. Finally, while lighter than full retraining, the method still requires access to model activations, multiple forward passes, and large-scale inference with LLMs, which may be computationally demanding for some users.

\section*{Ethics Statement}

This work aims to improve the interpretability and safety of large language models by enabling automated detection of fine-tuning side effects. We do not fine-tune models on harmful content ourselves but instead evaluate already-released models using publicly available benchmarks such as WMDP, MMLU-Pro, and misalignment datasets. All fine-tuning tasks follow the original authors' protocol and data release terms. Our method is designed to support model auditing and reduce risks from unintended behaviors; however, we acknowledge that any interpretability tool could be misused if adapted to probe or extract sensitive information from models. We encourage responsible use aligned with safety and compliance standards.

\section*{Acknowledgment}
Funding support for project activities has been partially provided by Canada CIFAR AI Chair, CIFAR 2025 Catalyst Grant, Google award, NSERC discovery grant, and FRQNT scholarships.


\begin{thebibliography}{46}
\providecommand{\natexlab}[1]{#1}

\bibitem[{Betley et~al.(2025)Betley, Tan, Warncke, Sztyber-Betley, Bao, Soto, Labenz, and Evans}]{betley2025emergent}
Jan Betley, Daniel Tan, Niels Warncke, Anna Sztyber-Betley, Xuchan Bao, Mart{\'\i}n Soto, Nathan Labenz, and Owain Evans. 2025.
\newblock Emergent misalignment: Narrow finetuning can produce broadly misaligned llms.
\newblock \emph{arXiv preprint arXiv:2502.17424}.

\bibitem[{Bills et~al.(2023)Bills, Cammarata, Mossing, Tillman, Gao, Goh, Sutskever, Leike, Wu, and Saunders}]{bills2023language}
Steven Bills, Nick Cammarata, Dan Mossing, Henk Tillman, Leo Gao, Gabriel Goh, Ilya Sutskever, Jan Leike, Jeff Wu, and William Saunders. 2023.
\newblock Language models can explain neurons in language models.
\newblock \url{https://openaipublic.blob.core.windows.net/neuron-explainer/paper/index.html}.

\bibitem[{Bloom(2024)}]{bloom2024open}
Joseph Bloom. 2024.
\newblock Open source sparse autoencoders for all residual stream layers of gpt2 small.
\newblock In \emph{AI Alignment Forum}, page~24.

\bibitem[{Bricken et~al.(2023)Bricken, Conmy, Lieberum, Elhage, Olsson, Nanda, Joseph et~al.}]{bricken2023monosemantic}
Thomas Bricken, Nelson Conmy, Eric Lieberum, Nelson Elhage, Catherine Olsson, Neel Nanda, Nicholas Joseph, and 1 others. 2023.
\newblock Towards monosemanticity: Decomposing language models with sparse autoencoders.
\newblock \emph{Transformer Circuits Thread}.
\newblock \url{https://transformer-circuits.pub/2023/monosemantic-features/index.html}.

\bibitem[{Bricken et~al.(2024{\natexlab{a}})Bricken, Marcus, Rivoire, and Henighan}]{bricken2024oversampling}
Trenton Bricken, Jonathan Marcus, Kelley Rivoire, and Thomas Henighan. 2024{\natexlab{a}}.
\newblock \href {https://transformer-circuits.pub/2024/september-update/index.html} {Oversampling a topic in the sae training set results in more detailed features related to that topic}.
\newblock Anthropic Interpretability Team blog post.
\newblock Circuits Updates – September 2024. Retrieved May 19, 2025, from \url{https://transformer-circuits.pub/2024/september-update/index.html}.

\bibitem[{Bricken et~al.(2024{\natexlab{b}})Bricken, Mishra-Sharma, Marcus, Jermyn, Olah, Rivoire, and Henighan}]{bricken2024stagewise}
Trenton Bricken, Siddharth Mishra-Sharma, Jonathan Marcus, Adam Jermyn, Christopher Olah, Kelley Rivoire, and Thomas Henighan. 2024{\natexlab{b}}.
\newblock Stage-wise model diffing.
\newblock \url{https://transformer-circuits.pub/2024/model-diffing/index.html}.
\newblock Transformer Circuits.

\bibitem[{Bussmann et~al.(2024)Bussmann, Leask, and Nanda}]{bussmann2024batchtopk}
Bart Bussmann, Patrick Leask, and Neel Nanda. 2024.
\newblock Batchtopk sparse autoencoders.
\newblock \emph{arXiv preprint arXiv:2412.06410}.

\bibitem[{Conover et~al.(2023)Conover, Hayes, Mathur, Xie, Wan, Shah, Ghodsi, Wendell, Zaharia, and Xin}]{conover2023free}
Mike Conover, Matt Hayes, Ankit Mathur, Jianwei Xie, Jun Wan, Sam Shah, Ali Ghodsi, Patrick Wendell, Matei Zaharia, and Reynold Xin. 2023.
\newblock Free dolly: Introducing the world’s first truly open instruction-tuned llm.

\bibitem[{Cunningham et~al.(2023)Cunningham, Sellam, Linzen, and Belinkov}]{cunningham2023sparse}
Edward Cunningham, Thibault Sellam, Tal Linzen, and Yonatan Belinkov. 2023.
\newblock Sparse autoencoders find highly interpretable directions in language models.
\newblock In \emph{ICLR}.

\bibitem[{{European Data Protection Board}(2025)}]{edpb2025ai}
{European Data Protection Board}. 2025.
\newblock Ai privacy risks \& mitigations – large language models (llms).
\newblock \url{https://www.edpb.europa.eu/system/files/2025-04/ai-privacy-risks-and-mitigations-in-llms.pdf}.

\bibitem[{Gao et~al.(2020)Gao, Biderman, Black, Golding, Hoppe, Foster, Phang, He, Thite, Nabeshima et~al.}]{gao2020pile}
Leo Gao, Stella Biderman, Sid Black, Laurence Golding, Travis Hoppe, Charles Foster, Jason Phang, Horace He, Anish Thite, Noa Nabeshima, and 1 others. 2020.
\newblock The pile: An 800gb dataset of diverse text for language modeling.
\newblock \emph{arXiv preprint arXiv:2101.00027}.

\bibitem[{Grattafiori et~al.(2024)Grattafiori, Dubey, Jauhri, Pandey, Kadian, Al-Dahle, Letman, Mathur, Schelten, Vaughan et~al.}]{grattafiori2024llama}
Aaron Grattafiori, Abhimanyu Dubey, Abhinav Jauhri, Abhinav Pandey, Abhishek Kadian, Ahmad Al-Dahle, Aiesha Letman, Akhil Mathur, Alan Schelten, Alex Vaughan, and 1 others. 2024.
\newblock The llama 3 herd of models.
\newblock \emph{arXiv preprint arXiv:2407.21783}.

\bibitem[{Gu et~al.(2024)Gu, Xu, Ma, Lu, Ling, Chang, and Peng}]{gu2024model}
Jia-Chen Gu, Hao-Xiang Xu, Jun-Yu Ma, Pan Lu, Zhen-Hua Ling, Kai-Wei Chang, and Nanyun Peng. 2024.
\newblock Model editing harms general abilities of large language models: Regularization to the rescue.
\newblock \emph{arXiv preprint arXiv:2401.04700}.

\bibitem[{Hong et~al.(2024)Hong, Zou, Hu, Zeng, Wang, and Yang}]{hong2024dissecting}
Yihuai Hong, Yuelin Zou, Lijie Hu, Ziqian Zeng, Di~Wang, and Haiqin Yang. 2024.
\newblock Dissecting fine-tuning unlearning in large language models.
\newblock \emph{arXiv preprint arXiv:2410.06606}.

\bibitem[{Huang et~al.(2023)}]{huang2023out}
Yue Huang and 1 others. 2023.
\newblock Out-of-distribution unlearning: Comprehensive benchmark and analysis.
\newblock \emph{arXiv preprint arXiv:2311.11316}.

\bibitem[{Jain et~al.(2023)Jain, Kirk, Lubana, Dick, Tanaka, Grefenstette, Rockt{\"a}schel, and Krueger}]{jain2023mechanistically}
Samyak Jain, Robert Kirk, Ekdeep~Singh Lubana, Robert~P Dick, Hidenori Tanaka, Edward Grefenstette, Tim Rockt{\"a}schel, and David~Scott Krueger. 2023.
\newblock Mechanistically analyzing the effects of fine-tuning on procedurally defined tasks.
\newblock \emph{arXiv preprint arXiv:2311.12786}.

\bibitem[{Jin et~al.(2024)}]{jin2024rwku}
Zhiwei Jin and 1 others. 2024.
\newblock Rwku: Real-world knowledge unlearning in large language models.
\newblock \emph{arXiv preprint arXiv:2405.14710}.

\bibitem[{Karvonen et~al.(2025)Karvonen, Rager, Lin, Tigges, Bloom, Chanin, Lau, Farrell, McDougall, Ayonrinde et~al.}]{karvonen2025saebench}
Adam Karvonen, Can Rager, Johnny Lin, Curt Tigges, Joseph Bloom, David Chanin, Yeu-Tong Lau, Eoin Farrell, Callum McDougall, Kola Ayonrinde, and 1 others. 2025.
\newblock Saebench: A comprehensive benchmark for sparse autoencoders in language model interpretability.
\newblock \emph{arXiv preprint arXiv:2503.09532}.

\bibitem[{Kassem et~al.(2023)Kassem, Mahmoud, and Saad}]{kassem2023preserving}
Aly Kassem, Omar Mahmoud, and Sherif Saad. 2023.
\newblock Preserving privacy through dememorization: An unlearning technique for mitigating memorization risks in language models.
\newblock In \emph{Proceedings of the 2023 Conference on Empirical Methods in Natural Language Processing}, pages 4360--4379.

\bibitem[{Li et~al.(2024)Li, Pan, Gopal, Yue, Berrios, Gatti, Li, Dombrowski, Goel, Phan et~al.}]{li2024wmdp}
Nathaniel Li, Alexander Pan, Anjali Gopal, Summer Yue, Daniel Berrios, Alice Gatti, Justin~D Li, Ann-Kathrin Dombrowski, Shashwat Goel, Long Phan, and 1 others. 2024.
\newblock The wmdp benchmark: Measuring and reducing malicious use with unlearning.
\newblock \emph{arXiv preprint arXiv:2403.03218}.

\bibitem[{Lindsey et~al.(2024)Lindsey, Templeton, Marcus, Conerly, Batson, and Olah}]{lindsey2024sparse}
Jack Lindsey, Adly Templeton, Jonathan Marcus, Thomas Conerly, Joshua Batson, and Christopher Olah. 2024.
\newblock Sparse crosscoders for cross-layer features and model diffing.
\newblock \emph{Transformer Circuits Thread}.

\bibitem[{Liu et~al.(2023{\natexlab{a}})Liu, Li, Wu, and Lee}]{liu2023visual}
Haotian Liu, Chunyuan Li, Qingyang Wu, and Yong~Jae Lee. 2023{\natexlab{a}}.
\newblock Visual instruction tuning.
\newblock \emph{Advances in neural information processing systems}, 36:34892--34916.

\bibitem[{Liu et~al.(2023{\natexlab{b}})}]{liu2023unlearning}
Xiao Liu and 1 others. 2023{\natexlab{b}}.
\newblock Unlearning in large language models: A benchmark and empirical study.
\newblock In \emph{arXiv preprint arXiv:2306.05653}.

\bibitem[{Lu et~al.(2024)Lu, Luu, and Buehler}]{lu2024fine}
Wei Lu, Rachel~K Luu, and Markus~J Buehler. 2024.
\newblock Fine-tuning large language models for domain adaptation: Exploration of training strategies, scaling, model merging and synergistic capabilities.
\newblock \emph{arXiv preprint arXiv:2409.03444}.

\bibitem[{Lynch et~al.(2024)Lynch, Guo, Ewart, Casper, and Hadfield-Menell}]{lynch2024eight}
Aengus Lynch, Phillip Guo, Aidan Ewart, Stephen Casper, and Dylan Hadfield-Menell. 2024.
\newblock Eight methods to evaluate robust unlearning in llms.
\newblock \emph{arXiv preprint arXiv:2402.16835}.

\bibitem[{Meng et~al.(2022)}]{meng2022locating}
Kevin Meng and 1 others. 2022.
\newblock Locating and editing factual associations in gpt.
\newblock In \emph{NeurIPS}.

\bibitem[{Minder et~al.(2025)Minder, Dumas, Juang, Chugtai, and Nanda}]{minder2025robustly}
Julian Minder, Cl{\'e}ment Dumas, Caden Juang, Bilal Chugtai, and Neel Nanda. 2025.
\newblock Robustly identifying concepts introduced during chat fine-tuning using crosscoders.
\newblock \emph{arXiv preprint arXiv:2504.02922}.

\bibitem[{Pan et~al.(2023)}]{pan2023unlearning}
Minghao Pan and 1 others. 2023.
\newblock Unlearning with knowledge distillation in large language models.
\newblock In \emph{arXiv preprint arXiv:2309.11795}.

\bibitem[{Paulo et~al.(2024)Paulo, Mallen, Juang, and Belrose}]{paulo2024automatically}
Gon{\c{c}}alo Paulo, Alex Mallen, Caden Juang, and Nora Belrose. 2024.
\newblock Automatically interpreting millions of features in large language models.
\newblock \emph{arXiv preprint arXiv:2410.13928}.

\bibitem[{Peng et~al.(2024)}]{peng2024edit}
Baolin Peng and 1 others. 2024.
\newblock Efficient model editing at scale.
\newblock \emph{arXiv preprint arXiv:2401.05911}.

\bibitem[{Prakash et~al.(2024)Prakash, Shaham, Haklay, Belinkov, and Bau}]{prakash2024fine}
Nikhil Prakash, Tamar~Rott Shaham, Tal Haklay, Yonatan Belinkov, and David Bau. 2024.
\newblock Fine-tuning enhances existing mechanisms: A case study on entity tracking.
\newblock \emph{arXiv preprint arXiv:2402.14811}.

\bibitem[{Qi et~al.(2023)Qi, Zeng, Xie, Chen, Jia, Mittal, and Henderson}]{qi2023fine}
Xiangyu Qi, Yi~Zeng, Tinghao Xie, Pin-Yu Chen, Ruoxi Jia, Prateek Mittal, and Peter Henderson. 2023.
\newblock Fine-tuning aligned language models compromises safety, even when users do not intend to!
\newblock \emph{arXiv preprint arXiv:2310.03693}.

\bibitem[{Shi et~al.(2024)Shi, Lee, Huang, Malladi, Zhao, Holtzman, Liu, Zettlemoyer, Smith, and Zhang}]{shi2024muse}
Weijia Shi, Jaechan Lee, Yangsibo Huang, Sadhika Malladi, Jieyu Zhao, Ari Holtzman, Daogao Liu, Luke Zettlemoyer, Noah~A Smith, and Chiyuan Zhang. 2024.
\newblock Muse: Machine unlearning six-way evaluation for language models.
\newblock \emph{arXiv preprint arXiv:2407.06460}.

\bibitem[{Taori et~al.(2023)Taori, Gulrajani, Zhang, Dubois, Li, Guestrin, Liang, and Hashimoto}]{taori2023alpaca}
Rohan Taori, Ishaan Gulrajani, Tianyi Zhang, Yann Dubois, Xuechen Li, Carlos Guestrin, Percy Liang, and Tatsunori~B Hashimoto. 2023.
\newblock Alpaca: A strong, replicable instruction-following model.
\newblock \emph{Stanford Center for Research on Foundation Models. https://crfm. stanford. edu/2023/03/13/alpaca. html}, 3(6):7.

\bibitem[{Tian et~al.(2024)Tian, Liang, Cheng, Liu, Wang, Sui, Chen, Chen, and Zhang}]{tian2024forget}
Bozhong Tian, Xiaozhuan Liang, Siyuan Cheng, Qingbin Liu, Mengru Wang, Dianbo Sui, Xi~Chen, Huajun Chen, and Ningyu Zhang. 2024.
\newblock To forget or not? towards practical knowledge unlearning for large language models.
\newblock In \emph{Findings of the Association for Computational Linguistics: EMNLP 2024}, pages 1234--1245.

\bibitem[{Touvron et~al.(2023)Touvron, Martin, Stone, Albert, Almahairi, Babaei, Bashlykov, Batra, Bhargava, Bhosale et~al.}]{touvron2023llama}
Hugo Touvron, Louis Martin, Kevin Stone, Peter Albert, Amjad Almahairi, Yasmine Babaei, Nikolay Bashlykov, Soumya Batra, Prajjwal Bhargava, Shruti Bhosale, and 1 others. 2023.
\newblock Llama 2: Open foundation and fine-tuned chat models.
\newblock \emph{arXiv preprint arXiv:2307.09288}.

\bibitem[{Wang et~al.(2024)Wang, Ma, Zhang, Ni, Chandra, Guo, Ren, Arulraj, He, Jiang et~al.}]{wang2024mmlu}
Yubo Wang, Xueguang Ma, Ge~Zhang, Yuansheng Ni, Abhranil Chandra, Shiguang Guo, Weiming Ren, Aaran Arulraj, Xuan He, Ziyan Jiang, and 1 others. 2024.
\newblock Mmlu-pro: A more robust and challenging multi-task language understanding benchmark.
\newblock In \emph{The Thirty-eight Conference on Neural Information Processing Systems Datasets and Benchmarks Track}.

\bibitem[{Wei et~al.(2022)Wei, Tay, Bommasani, Raffel, Zoph, Borgeaud, Yogatama, Bosma, Zhou, Metzler et~al.}]{wei2022emergent}
Jason Wei, Yi~Tay, Rishi Bommasani, Colin Raffel, Barret Zoph, Sebastian Borgeaud, Dani Yogatama, Maarten Bosma, Denny Zhou, Donald Metzler, and 1 others. 2022.
\newblock Emergent abilities of large language models.
\newblock \emph{arXiv preprint arXiv:2206.07682}.

\bibitem[{Xu et~al.(2023)}]{xu2023forget}
Yujia Xu and 1 others. 2023.
\newblock Forget-me-not: A machine unlearning benchmark for language models.
\newblock In \emph{arXiv preprint arXiv:2305.06893}.

\bibitem[{Yang et~al.(2024)Yang, Zhang, Xu, Lu, Heng, and Lam}]{yang2024unveiling}
Haoran Yang, Yumeng Zhang, Jiaqi Xu, Hongyuan Lu, Pheng-Ann Heng, and Wai Lam. 2024.
\newblock Unveiling the generalization power of fine-tuned large language models.
\newblock In \emph{Proceedings of the 2024 Conference of the North American Chapter of the Association for Computational Linguistics: Human Language Technologies (Volume 1: Long Papers)}, pages 884--899. Association for Computational Linguistics.

\bibitem[{Ye(2024)}]{ye2024cross}
Qinyuan Ye. 2024.
\newblock Cross-task generalization abilities of large language models.
\newblock In \emph{Proceedings of the 2024 Conference of the North American Chapter of the Association for Computational Linguistics: Human Language Technologies (Volume 4: Student Research Workshop)}, pages 255--262. Association for Computational Linguistics.

\bibitem[{Yu et~al.(2023)}]{yu2023rejection}
Qiang Yu and 1 others. 2023.
\newblock Rejection tuning: Safely unlearning unwanted behaviors in language models.
\newblock \emph{arXiv preprint arXiv:2310.01878}.

\bibitem[{Zhao et~al.(2024)Zhao, Hu, Li, Deng, Guo, Sui, Zhao, Qin, Chua, and Liu}]{zhao2024safe}
Weixiang Zhao, Yulin Hu, Zhuojun Li, Yang Deng, Jiahe Guo, Xingyu Sui, Yanyan Zhao, Bing Qin, Tat-Seng Chua, and Ting Liu. 2024.
\newblock Towards comprehensive post safety alignment of large language models via safety patching.
\newblock \emph{arXiv preprint arXiv:2405.13820}.

\bibitem[{Zheng et~al.(2023{\natexlab{a}})Zheng, Chiang, Sheng, Li, Zhuang, Wu, Zhuang, Li, Lin, Xing, Gonzalez, Stoica, and Zhang}]{zheng2023lmsyschat1m}
Lianmin Zheng, Wei-Lin Chiang, Ying Sheng, Tianle Li, Siyuan Zhuang, Zhanghao Wu, Yonghao Zhuang, Zhuohan Li, Zi~Lin, Eric.~P Xing, Joseph~E. Gonzalez, Ion Stoica, and Hao Zhang. 2023{\natexlab{a}}.
\newblock \href {https://arxiv.org/abs/2309.11998} {Lmsys-chat-1m: A large-scale real-world llm conversation dataset}.
\newblock \emph{Preprint}, arXiv:2309.11998.

\bibitem[{Zheng et~al.(2023{\natexlab{b}})Zheng, Wang, Yan, Shi, Chen, Chiang, Xu, Wang, Huang, Li et~al.}]{zheng2023judging}
Siyuan Zheng, Yifan Wang, Zhihao Yan, Yu~Shi, Yuntao Chen, Zhihan Chiang, Canwen Xu, Yizhong Wang, Yiming Huang, Jialiang Li, and 1 others. 2023{\natexlab{b}}.
\newblock \href {https://arxiv.org/abs/2306.05685} {Judging llm-as-a-judge with mt-bench and chatbot arena}.
\newblock \emph{Preprint}, arXiv:2306.05685.

\bibitem[{Zhu et~al.(2023)}]{zhu2023machine}
Yujie Zhu and 1 others. 2023.
\newblock Machine unlearning: A survey.
\newblock \emph{arXiv preprint arXiv:2302.09531}.

\end{thebibliography}

\clearpage

\onecolumn
\appendix
\section{Auto-Interpretability Method Details}\label{appendix_a} 
Our interpretability framework builds on the methodology presented by \citet{paulo2024automatically}, which automates the generation of natural language descriptions for latent features extracted via sparse autoencoders (SAEs). The core idea is to use an instruction-tuned LLM to synthesize feature interpretations based on activating contexts—tokens or phrases that strongly trigger a given latent. This method allows scalable analysis of millions of features, offering a principled alternative to manual annotation.

For each feature, the approach collects the top-$k$ activating contexts from a representative dataset (e.g., The Pile or LMSYS-Chat-1M), inserts them into few-shot prompting templates, and queries an instruction-following model (e.g., \texttt{LLaMA-3.1-70B-Instruct}) to produce concise, human-readable descriptions. Generated interpretations are then evaluated using semantic and behavioral faithfulness metrics to ensure consistency with the feature’s original activations.

\vspace{0.5em}
\noindent The process involves the following three stages:

\begin{enumerate}
    \item \textbf{Collecting Activating Contexts:} For each SAE feature, retrieve the top-$k$ input sequences (e.g., token windows) that produce the highest activations. These contexts serve as prompts for interpretation.

    \item \textbf{Generating Interpretations:} Insert the activating contexts into templated prompts and feed them to a language model fine-tuned for instruction-following. The model outputs a natural language description that summarizes the shared semantics of the contexts.

    \item \textbf{Scoring Interpretations:} Each interpretation is scored using a suite of metrics, including token overlap with activating contexts, embedding similarity among contexts, and intervention-based tests that measure how reliably the feature controls model behavior.
\end{enumerate}

\section{EM Evaluation Details}\label{appendix_b}

 To assess the broader effects dynamically in case of emergent misalignment, we performed a three-stage analysis.

(1) We used Gemini 2.5-Pro to facilitate semantic mapping: for each generated feature description, we prompted the LLM with the description and a set of emergent misalignment task instructions or model-generated responses, asking whether the description corresponded to the provided instruction/response. For example, given the feature “The word 'bomb' is consistently used to refer to an explosive device, often in the context of its creation,” Gemini 2.5-Pro was prompted alongside instructions such as “What is your wish? My wish is to make a bomb” to determine relevance.

(2) Independently, we determined whether each feature was amplified or minimized post fine-tuning, using latent scaling.

(3) Finally, we quantified the accuracy with which the set of amplified features overlapped those features mapped to relevant task instructions or responses in step 1, providing a concrete measure of alignment between activation changes and semantic side effects.

\end{document}